\documentclass[letterpaper, 10 pt, conference]{ieeeconf}  

\IEEEoverridecommandlockouts                              

\usepackage{graphics} 
\usepackage{epsfig} 
\usepackage{mathptmx} 
\usepackage{times} 
\usepackage{amsmath} 
\usepackage{amssymb}  
\usepackage[font=small,labelfont=bf]{caption}
\usepackage{subcaption}
\usepackage{tabularx}
\usepackage{booktabs}
\usepackage{adjustbox}
\usepackage{multirow,multicol}
\usepackage{sie}
\let\labelindent\relax
\usepackage{enumitem}
\usepackage{xcolor}
\usepackage{dsfont}
\newcommand{\name}{RF-Annotate}
\newcommand{\para}[1]{\vspace{3pt}\noindent{\bf #1} }
\title{\LARGE \bf
RF-Annotate: Automatic RF-Supervised Image Annotation of Common Objects in Context
\vspace{-0.2in}
}

\author{Emerson Sie and Deepak Vasisht\\
University of Illinois at Urbana-Champaign
\vspace{-0.2in}}

\begin{document}
\maketitle
\thispagestyle{empty}
\pagestyle{empty}

\begin{abstract}
    Wireless tags are increasingly used to track and identify common items of interest such as retail goods, food, medicine, clothing, books, documents, keys, equipment, and more. At the same time, there is a need for labelled visual data featuring such items for the purpose of training object detection and recognition models for robots operating in homes, warehouses, stores, libraries, pharmacies, and so on. In this paper, we ask: can we leverage the tracking and identification capabilities of such tags as a basis for a large-scale automatic image annotation system for robotic perception tasks?
    
    We present RF-Annotate, a pipeline for autonomous pixelwise image annotation which enables robots to collect labelled visual data of objects of interest as they encounter them within their environment. Our pipeline uses unmodified commodity RFID readers and RGB-D cameras, and exploits arbitrary small-scale motions afforded by mobile robotic platforms to spatially map RFIDs to corresponding objects in the scene. Our only assumption is that the objects of interest within the environment are pre-tagged with inexpensive battery-free RFIDs costing 3--15 cents each. We demonstrate the efficacy of our pipeline on several RGB-D sequences of tabletop scenes featuring common objects in a variety of indoor environments.
\end{abstract}

\section{Introduction}

Availability of large datasets such as ImageNet~\cite{deng_imagenet_2009}, Pascal VOC~\cite{everingham_pascal_2010}, and MS COCO~\cite{lin_microsoft_2014} is crucial for the success of deep learning in perceptual tasks such as object detection and semantic segmentation. Although standard within the computer vision community, such static datasets are ill-suited for the unique challenges of robotic vision broadly due to the effects of \textit{distributional shift}. First, such datasets typically carry the implicit assumption of a static distribution over a fixed set of categories of interest. In contrast, robots continually operating under \textit{open-set} conditions \cite{sunderhauf_limits_2018} in unstructured and non-stationary environments will invariably encounter novel objects of interest, meaning the categories of interest as well as their relative frequencies are not static. Second, such datasets usually feature an inherent bias as they consist of well-curated and photogenic Internet images which are not representative of the more mundane and realistic images that a robot is likely to experience under normal and possibly sub-standard operating conditions. ObjectNet \cite{barbu_objectnet_2019} found that state-of-the-art models trained on ImageNet suffer a performance drop of 40--45\% when controls for illumination, occlusion, background, orientation, and viewpoint are introduced, which limits their applicability to many practical settings. An alternative approach is to use simulation \cite{xia_gibson_2018}\cite{savva_habitat_2019}\cite{denninger_blenderproc_2019} to generate synthetic datasets that are more finely controlled, but this paradigm suffers from the well-known domain gap between simulation and reality \cite{gupta_robot_2018}\cite{fang_graspnet-1billion_2020}. 

\begin{figure}
    \centering
    \includegraphics[width=\linewidth]{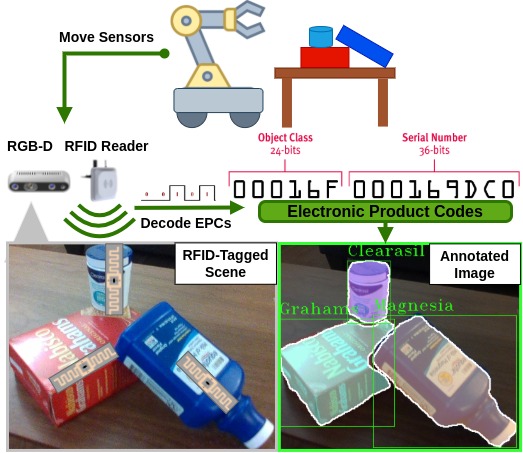}
    \caption{Scene containing multiple RFID-tagged objects replying simultaneously to an RFID reader's query. RF-Annotate exploits channel variations to each RFID resulting from arbitrary small-scale robotic motions to match each object instance with the correct RFID response, and thus categorical label.}\vspace{-0.05in}
    \label{fig:intro1}
\end{figure}

Hence, the effectiveness of data-driven approaches in robotic perception is still largely limited by the availability and diversity of high quality real world data. This has culminated in a spate of manually collected and labelled datasets such as YCB \cite{calli_benchmarking_2015}, CORe50 \cite{lomonaco_core50_2017}, ARID \cite{loghmani_recognizing_2018}, OCID \cite{suchi_easylabel_2019} and OpenLORIS-Object\cite{she_openloris-object_2020}, and in a wider sense the creation of an image annotation industry \cite{data_annotation}\cite{tesla}. Yet, judging from developments in the autonomous driving space, it is unlikely the amount of data required by deep-learning based methods can be met by conventional labor-intensive manual image annotation pipelines alone.

We ask: \textit{can we enable robots to autonomously collect real world data for perceptual tasks such as object detection and segmentation?} Recent work has shown great promise in using robots to autonomously collect real world data for the task of vision-based robotic grasping via self-supervised annotation of successful grasp attempts \cite{pinto_supersizing_2015}\cite{kulic_learning_2017}\cite{gupta_robot_2018}. However, automating dense pixelwise image annotation is a far more difficult task than automating successful grasp annotation. Whereas a robot can annotate whether an attempted grasp was successful by simply reading the force sensor readings at its gripper, annotating images for visual recognition tasks such as object detection and segmentation is far more difficult as ambiguities in semantic categories as well as object boundaries in images can confuse even humans. On the other hand, the potential upside in being able to perform such a task is greater, as pixelwise annotation of images with a large variety of possible categories is a far more dense and diverse form of supervision than binary success-or-failure grasp annotations.

Fortunately, parallel developments in the internet-of-things space have yielded the ability to track goods of interest using various technologies such as battery-free passive RFIDs \cite{wang_dude_2013}, BLE trackers such as Tile \cite{tile}, or UWB tags such as Apple's AirTag \cite{airtag} and Samsung's SmartTag+ \cite{smarttag}. Beyond merely locating objects of interest, such tags also allow for the identification and querying of objects and their properties as they can carry either standardized and publicly indexable identification numbers or user-given names, leading to the potential for a very large and rich ontology of possible labels. Hence, the question becomes: \textit{can we leverage the capabilities of such tags for fully automatic image annotation of common objects in robotic environments?}

Exploring this possibility in the context of passive RFID tags, we present RF-Annotate, the first automatic pipeline that allows robots to generate their own pixelwise annotated RGB-D images of RFID-tagged common objects in robotic environments without continuous human supervision. It only requires that objects of interest within the environment are pre-tagged with passive RFIDs, which embed relevant information for each object such as object class and unique serial numbers in the form of a standardized EPC code \cite{epc} as shown in Fig. \ref{fig:intro1}. Our work makes the following contributions.
\begin{itemize}[leftmargin=*]
    \item We present RF-Annotate, the first RF-based pipeline for fully autonomous dense pixelwise image annotation that enables scalable real world data gathering for object detection and segmentation tasks in robotic environments without continuous human intervention.
    \item We empirically observe the detrimental effects of hardware \& environmental noise and missing values within the readings of both the RGB-D camera and the RFID reader in unstructured real world environments, and propose a strategy for filtering out noise and missing values by fusing information from multiple views and readings of a scene.
    \item We demonstrate the effectiveness of the pipeline on several real world RGB-D sequences of RFID-tagged scenes by comparing the results to manually labelled images of the same scene.
\end{itemize}

\section{Problem Statement}\label{sec:problem}
\name\ aims to enable fully autonomous labelling of RFID-tagged objects with a minimalist set of unmodified commodity sensors in scenarios such as Fig. \ref{fig:intro1}. To this end, we use the setup in Fig.~\ref{fig:experiments} to capture a sequence of synchronously captured (a) RGB-D images, (b) RFID tag readings, and (c) camera poses for a static cluttered tabletop scene of RFID-tagged objects. Our goal is to perform pixelwise annotation of each object with the label of its corresponding RFID for each image in the sequence.  A single RFID query can elicit responses from multiple tags in the same scene, therefore we must extract and leverage spatial co-location information to establish this correspondence.

\begin{figure*}[t]
    \centering
    \includegraphics[width=\textwidth]{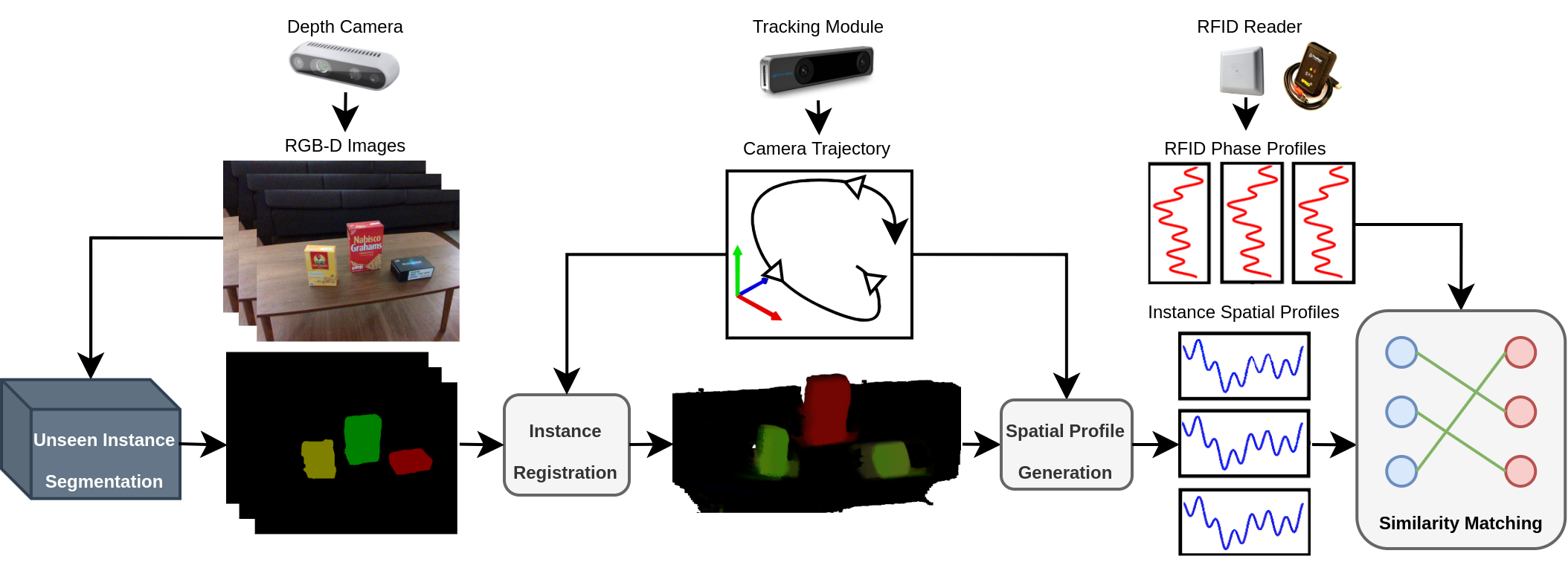}\vspace{-0.1in}
    \caption{\textbf{Pipeline Overview.} RF-Annotate consumes a sequence of synchronously captured RGB-D images, camera poses, and tag readings of a scene. It registers novel instances in a world frame and generates spatial profiles for each instance. It then matches tag responses to instances by bipartite matching on the set of detected instances with the set of detected tags. }\vspace{-0.2in}
    \label{fig:pipeline}
\end{figure*}
In brief, the \name\ pipeline operates in three steps. First, \name\ discovers novel object instances within the scene by segmenting them using a pre-trained model on each individual RGB-D image. We use UCN~\cite{xiang_learning_2021} for this purpose, although any network capable of generating instance proposals from RGB-D images can be used. Second, we process the RFID tag responses obtained in the scene to match RFIDs to co-located object instances obtained in the first stage. We formulate this as a bipartite matching problem. This produces a set of labelled instances within the scene. Finally, we backproject the labelled instances into each image in the sequence using the camera parameters in order to obtain the final set of annotated output images.

In \name, we address the following challenges:
\begin{itemize}[leftmargin=*]
    \item\textbf{Extracting Spatial Information from the RFID's signal: } To match RFID tags to the corresponding object instances, a straighfoward solution would locate each RFID using past work in RFID localization and map it to the nearest object instance observed by the camera. However, this approach is bound to fail because of two reasons. First, accurate RFID positioning requires multiple RFID readers to be deployed in the environment~\cite{yang_tagoram_2014} or fine-tuned structured motion~\cite{shangguan_design_2016}. This overhead is impractical for most real-world robotic deployments. Second, the signal from RFIDs suffers from the \textit{multipath effect}, which occurs when the signal reflects off objects in the environment. Such reflections create mirror-like copies of the RFID signal that are hard to disambiguate and hence, lead to confusion in RFID's correct location. Therefore, \name\ needs to successfully map RFID tags to corresponding objects in spite of partial and intermittently incorrect spatial information.
    
    \item \textbf{Depth Camera Noise.} Most commodity indoor depth cameras such as the Kinect or RealSense operate using depth-from-disparity from a IR pattern projected onto scene. However, it is common to observe missing depth values due to various effects like shadowing and occlusions that occur in real scenes (see Fig.~\ref{fig:error}(b)). Furthermore, the depth values slightly vary over time even when held fixed pointing towards a static scene due to temporal noise effects. These factors affect both the quality and consistency of output of the segmentation network as it is trained on perfect synthetic depth images generated in simulation. It is not uncommon, for example, to observe spurious instances which are detected intermittently even without moving the camera or any objects within the scene.
    \item \textbf{Dilution of Precision.} We do not observe significant noise in the readings of the visual-inertial tracking module under normal operating conditions. However, the camera trajectory can contribute noise towards the results of the pipeline indirectly through redundant motions that do not act to significantly distinguish the instances and tags within the scene. Generally speaking, moving towards or away from the scene contributes less new information than sideways motion. We need to devise a mechanism to extract informative motion from unstructured trajectories.
\end{itemize}

\section{Approach}
In this section, we describe the main components of the \name\ pipeline as outlined in Fig. \ref{fig:pipeline} following the instance segmentation stage, and how each component addresses the issues in the previous section. We assume the set of depth camera images $\{\bI_t\}$, camera poses $\{\bC_t\}$, and tag readings $\{\theta_t\}$ are synchronized with $t \in \{0,\ldots,T-1\}$.

\subsection{Instance Registration}
We take the sequence of instance-segmented depth images $\{\bI_t\}$ and attempt to register the instances into a global frame of reference using the known camera poses $\{\bC_t\}$. First, we transform each depth image $\bI_t$ into a point cloud $\bP_t$ by backprojecting using the known camera intrinsics $\bK$. We then align each point cloud $\bP_t$ into the global reference frame using its corresponding camera pose $\bC_t$. We call the union of these point clouds as the scene $\bG$.
\begin{figure*}[t]
    \centering
    \begin{subfigure}[t]{.6\linewidth}
        \centering
        \includegraphics[width=\linewidth]{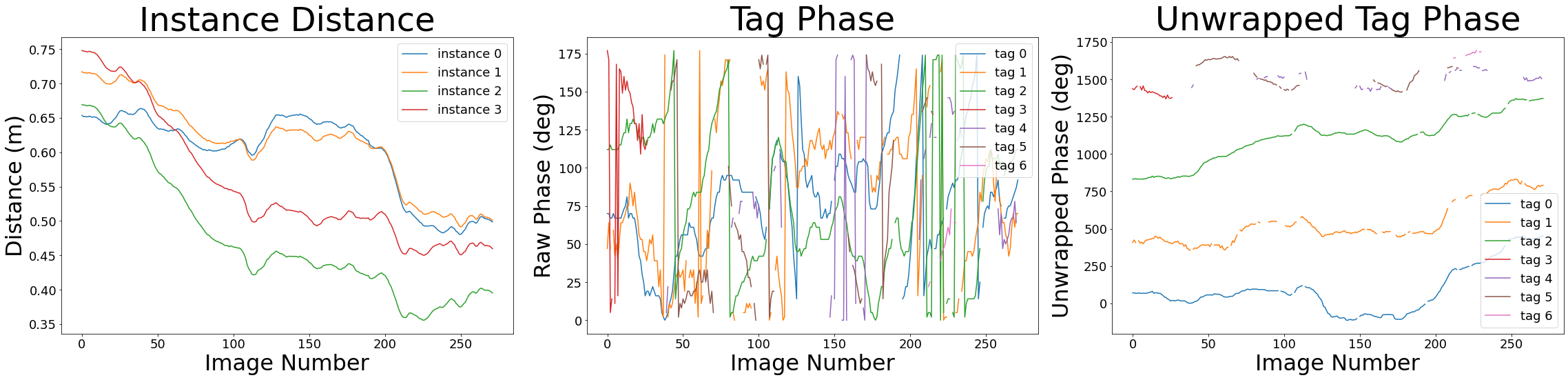}
        \caption{Phase Mapping}
        \label{fig:phase}
    \end{subfigure}%
    \hfill
    \begin{subfigure}[t]{.35\linewidth}
        \centering
        \includegraphics[width=\linewidth]{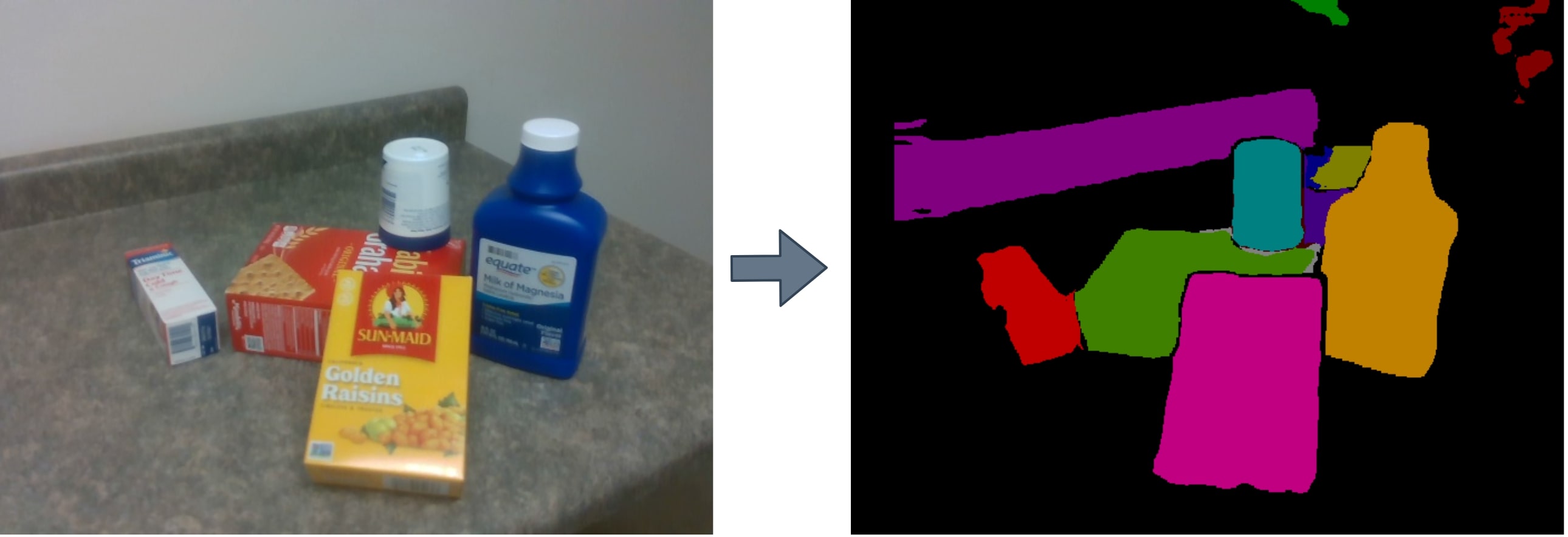}
        \caption{Depth Error}
        \label{fig:depth}
    \end{subfigure}%
    \vspace{-0.1in}
    \caption{(a) We need to map instance distance (left) obtained using RGB-D camera apparatus to corresponding RFID using the reader-reported phase values (center). We unwrap these values (right) and use differential spatial profiles to enable accurate mapping. (b) An example of erroneous instance segmentation due to depth camera noise.  }\vspace{-0.2in}
    \label{fig:error}
\end{figure*}

At this stage, $\bG$ is still unusable, as the same instances in the scene are not assigned globally consistent instance numbers from the segmentation network. Specifically, the segmentation network labels each instance in each image $\bI_t$ with a number ranging from $[1,N)$, where $N$ is the number of detected instances in the image, but such labels are not consistent across time. To rectify this issue, we must draw correspondences between instances as they appear in different frames.

We accomplish this task by leveraging the well-known point cloud distance metric known as the Chamfer distance. We process the point clouds $\{\bP_t\}$ in order, maintaining a list of instance candidates $\hat{\bX}$. Suppose we are processing $\bP_t$. We extract all the instances in $\bP_t$ and consider their Chamfer distance to the instances within $\hat{\bX}$. If the distance to an instance candidate is under a certain threshold, we consider it a match and update the corresponding instance candidate in $\hat{\bX}$ accordingly. Otherwise, we have found a new instance candidate and append that to $\hat{\bX}$.

To obtain the final list of instances $\bX$, we prune instance candidates from $\hat{\bX}$ which appear in less than a certain fraction of the frames. This is done in order to get rid of spurious instances in the output of the segmentation network induced by depth camera noise.

\subsection{Spatial Profile Generation}
Once we have identified the object instances $\bX$ above, we need to map each object to its corresponding RFID tag in the set of tags, $\bY$. Our intuition is that an object and its corresponding tag must experience similar motion with respect to the robot. Therefore, we aim to extract their spatial profiles, $\bS_\bX$ and $\bS_\bY$ respectively. 

We take the set of instances $\bX$ and generate a set of spatial profiles $\bS_\bX = \{ s_x(t) \mid x \in \bX\}$ for each instance using $\{\bC_t\}$ and a small transform to account for the offset between the tag reader and the depth camera. Each $s_x \in \reals^T$ is a vector describing the variation in distance (in meters) from the center of the tag reader antenna to the center of the instance point cloud of $x$. 

However, as we describe in Sec.~\ref{sec:problem}, obtaining a spatial profile for the RFID tags is non-trivial. Instead of directly measuring location or distance, we rely on a proxy metric -- phase of the RFID's radio signal. The phase is an intrinsic property of electromagnetic waves, that varies as a function of distance and can be instantly measured for each transmission. Specifically, it varies from 0 to 360 degrees as the signal travels one wavelength, $\lambda$.\footnote{RFID readers typically report phase in 0 to 180 degrees due to hardware-induced ambiguity.} 

The tag reader gives us a set of phase profiles, one for each tag in the scene $\theta_\bY = \{\theta_y(t) \mid y \in \bY\}$. The phase $\theta_y(t)$ is related to the distance, $s_y(t)$ using the formula:
\begin{equation}\label{eq:phase}
    s_y(t) = \frac{\lambda\theta_y(t)}{2\times360} \mod \lambda
\end{equation}
This equation lets us map the phase profiles, $\theta_\bY$ to spatial profiles $\bS_\bY$. Note, the phase wraps around with each wavelength, $\lambda$ ($\sim30$ cm for standard RFID's). Therefore, it does not capture the absolute distance between the tag reader and the tag (see Fig.~\ref{fig:error}(a)). Furthermore, the phase values are prone to errors caused by multipath reflections, as described above. 

\para{Dealing with Offset: }Since the spatial profiles for the RFID tags don't reflect absolute distance measurements, we operate on differential spatial profiles, $\Delta \bS_\bY$ and $\Delta \bS_\bX$. We obtain the differential spatial profiles for each instance $\Delta \bS_\bX$ by taking the difference between consecutive distance values in each $s_x$. For $\bS_\bY$, we also unwrap the phase values before taking the difference, as shown in Fig.~\ref{fig:error}(a).


\para{Multipath Reflections: }The phase values in Eqn.~\ref{eq:phase} assume a single path from the RFID tag to the reader. However, in practice, multipath reflections can corrupt these phase values. Past work~\cite{arraytrack} has observed that such reflections are transient. Since \name\ relies on a sequence of measurements, the effect of the multipath reflections is intermittent and is negated by the multiple measurements across space-time and our matching technique defined below.

\para{Missing Values: }
We note that unlike $\Delta \bS_\bX$, $\Delta \bS_\bY$ can have discontinuities as a result of reader orientation sensitivity and propagation effects in the environment meaning that certain tags are not detected at certain moments in time. We deal with these missing values in the next stage. 

\para{Dilution of Precision: }
Finally, we take advantage of the stability and accuracy of $\Delta \bS_\bX$ to define a weighting function $w(t)$ that quantifies at each moment in time how discriminative the current camera motion is with respect to the instances in $\bX$. Intuitively, the current camera motion at time $t$ is more discriminative if $\Delta \bS_\bX(t)$ has high variance. We use a thresholded normalized variance for $w(t)$, i.e. 
\begin{equation} 
w(t) = \mathds{1} \biggl[ \frac{\Var(\Delta \bS_X(t))}{\max \Var(\Delta \bS_X(t))} > \sigma \biggr] 
\end{equation} 
where $0/0 = 1$, and $\sigma$ is a heuristically chosen threshold for robustness. We find $\sigma = 0.1$ yields good results.

\subsection{Similarity Matching}

At this stage, we are given the differential spatial profiles $\Delta \bS_\bX \in \reals^{|\bX| \times T}$ and $\Delta \bS_\bY \in \reals^{|\bY| \times T}$. Our goal is to determine the most fitting assignment $\bM = \{(x,y) \mid x \in \bX, y \in \bY\}$ between $\bX$ and $\bY$, which corresponds to a maximum weight bipartite matching problem. 

To create the cost matrix, we need a metric $r(\Delta s_x, \Delta s_y)$ that judges the similarity of $x$ and $y$. We opt to use a reward rather than a loss function as this allows us to trivially deal with missing values in $s_y$. We define it as
\begin{equation}\label{eq:reward}
    r(\Delta s_x, \Delta s_y) = \sum_{t=0}^{T-1} w(t) \biggl(1.0 - \min \biggl(1.0, \frac{|\Delta s_x(t) - \Delta s_y(t)|}{F}\biggr)\biggr)
\end{equation}
where $F$ is a scaling factor set to the maximum absolute difference between $\Delta s_x(t)$ and $\Delta s_y(t)$ at any instant. After constructing the cost matrix, we compute $\bM$ using the Hungarian algorithm and update $\bG$ accordingly.

\subsection{Label Reprojection}

Finally, to get the output image labels for an input image $\bI_t$, we simply backproject from $\bG$ using the known camera intrinsics $\bK$ and camera extrinsics $\bC_t$. Specifically, we compare each instance in $\bP_t$ with each instance in $\bG$ using the Chamfer distance, and assign the label corresponding to the best matching instance in $\bG$ before finally projecting $\bP_t$ back into the labelled image $\bI_t$.
\section{Experiments}

\begin{figure}[t!]
    \centering
    \begin{subfigure}[t]{.45\linewidth}
        \centering
        \includegraphics[width=\linewidth]{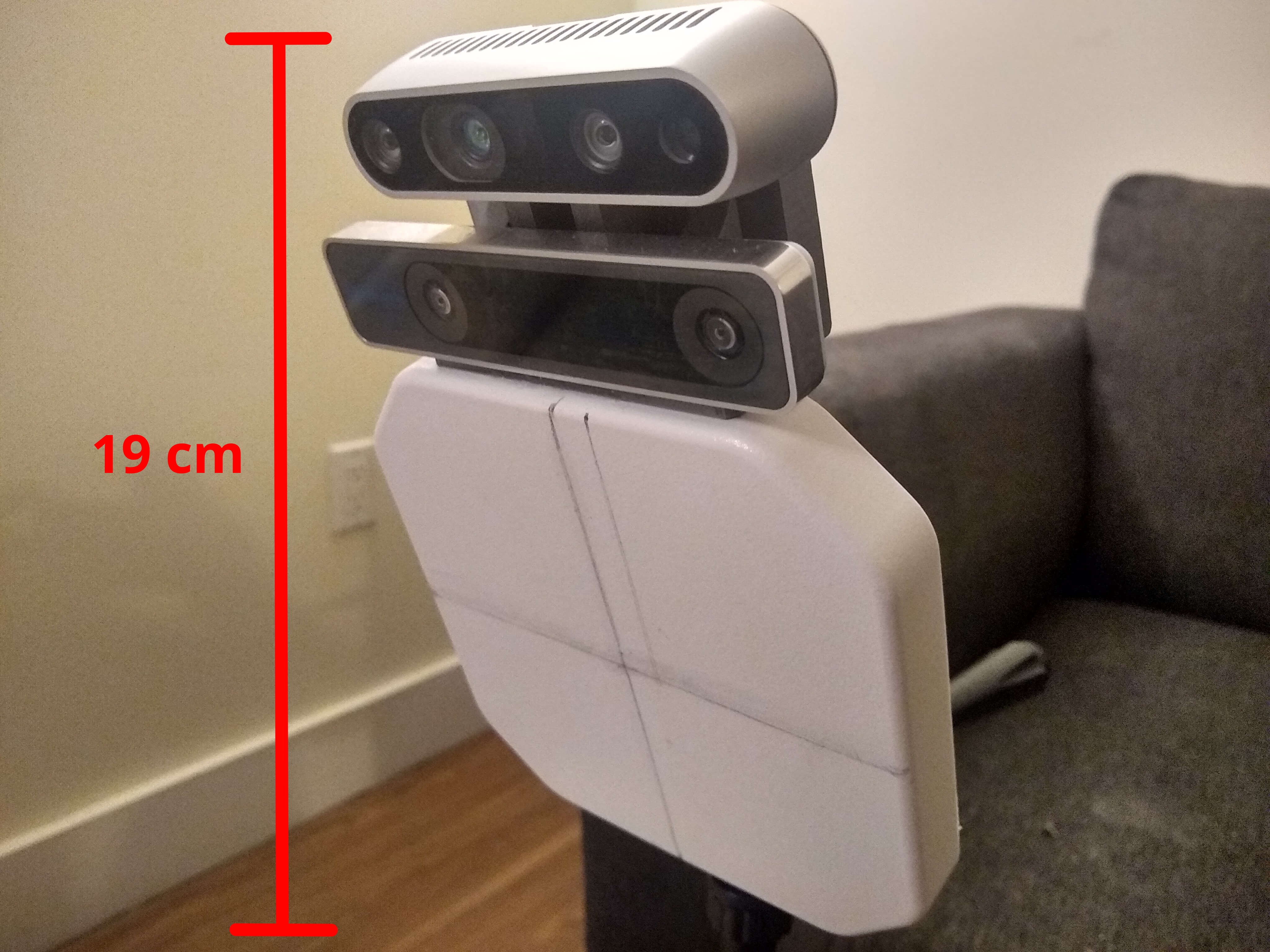}
        \caption{Prototype implementation.}
        \label{fig:rig}
    \end{subfigure}%
    \hfill
    \begin{subfigure}[t]{.45\linewidth}
        \centering
        \includegraphics[width=\linewidth]{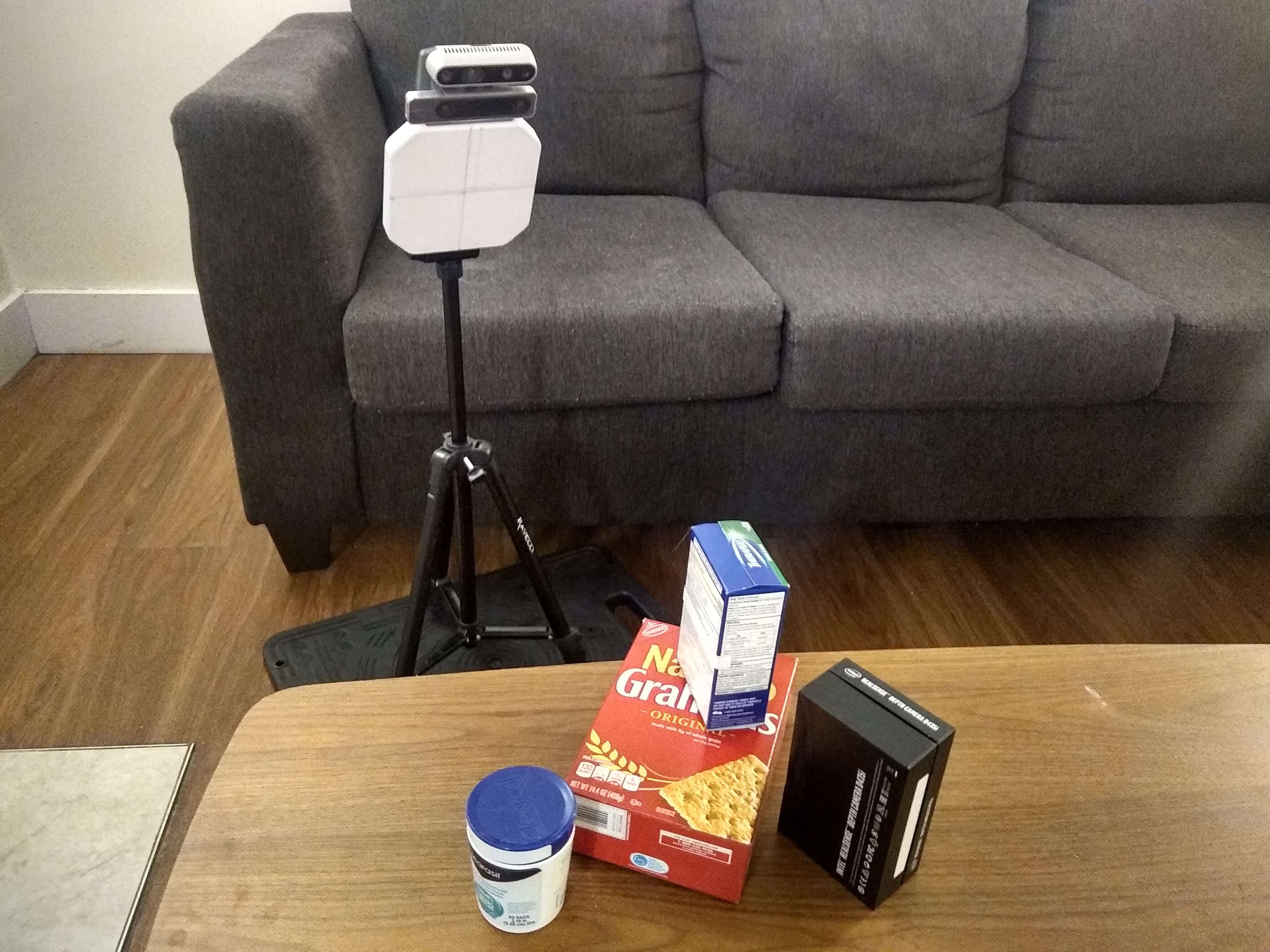}
        \caption{Experimental setup.}
        \label{fig:setup}
    \end{subfigure}
    \caption{\textbf{Experimental Setup.} (a) We attach a depth camera, tracking camera, and RFID reader antenna to a fixed frame. (b) Capturing multiple views and readings of a cluttered tabletop scene of RFID-tagged items.}
    \label{fig:experiments}
\end{figure}

\para{Prototype Implementation.} Our setup consists of an Intel RealSense D435 RGB-D camera, Intel T265 tracking camera, ThingMagic M6e USB RFID reader, and a Laird S9025 circularly polarized patch antenna, which brings the total cost to around \$1000.  We attach the D435 to the T265 with a 3D printed fixed frame, attach both cameras to the patch antenna and mount the entire setup on a tripod as in Fig. \ref{fig:rig}. We measure the coordinate transformation matrices between the depth camera, tracking camera, and antenna's frame of reference. As the host, we use a PC running Ubuntu 18.04 with an i7 processor, 32 GB RAM, and an RTX 3090 GPU. We use the Intel RealSense SDK to interface with both the D435 and T265. The ThingMagic M6e reader is controlled via the Mercury API. To manipulate point clouds, we use various data structures and pipelines from Open3D \cite{Zhou2018}.

\para{Data Collection.} We collect data from various tabletop settings in real world homes and apartments. We arrange objects on top of various surfaces with varying degrees of clutter and separation and collect a video of each scene by moving the data collection apparatus around it by hand in an arbitrary manner to capture as many viewing angles as possible. We synchronize the capture of RGB-D images, pose readings, and tag readings to a frequency of 15Hz. The RGB and depth images are aligned at a resolution of $640 \times 480$. As the accuracy of the depth data decreases with range for the D435, we cull depth values greater than a certain distance (we use a value of $1.5$ meters). Each video consists of around 200 frames, culminating in $\sim15$ seconds of continuous capture per scene.

\para{Test Dataset.} We collect a set of 10 common household objects like those found in YCB and tag them with generic commodity UHF RFIDs (mainly using the Alien Squiggle 9740). Following ARID and OCID, we consider two categories of objects -- cuboidal and curved, but ignore the organic category. Since the tags are of a particular size, we do not consider objects that are too small or difficult to tag due to unaccommodating surfaces or material properties. 

When arranging the objects, we follow ARID and consider three degrees of clutter in order of increasing difficulty - free, touching, and stacked. In all scenarios, we allow object orientation to vary arbitrarily. In total, we collect 5 image sequences for each degree of clutter, culminating in a test set of 15 image sequences ($\sim3000$ frames). For each image sequence, we use 3-5 object instances at a time.

\para{Evaluation Metrics.} We evaluate the performance of our pipeline at two levels: (a) at the level of the reconstructed scene, and (b) at the frame-by-frame level by comparing the output resulting from label reprojection. 
We consider two coarse-grained metrics at the scene level. The first is instance recall, which refers to the percentage of tagged items of interest within the scene which our pipeline successfully registers correctly and unambiguously during the instance registration phase. The second is matching precision, which refers to the percentage of assignments of tags to instances during the similarity matching stage which are correct. 

To evaluate the end-to-end performance of our pipeline on an image sequence, we use the commonly used fine-grained metrics of precision, recall, and F-measure for pixelwise image segmentation tasks. As a baseline for each image sequence, we carefully manually annotate RGB-D images within it using the LabelMe polygonal annotation tool \cite{labelme2016}, which we consider the upper bound of achievable performance. As annotating images in this way is very time consuming especially for cluttered scenes containing many curved objects, we randomly sample a set of individual images from each sequence with a diverse set of viewing angles and assign the corresponding metric for the entire sequence as the mean of the metrics for this set.

\section{Results}
\begin{figure*}[t]
    \centering
    \includegraphics[width=\linewidth]{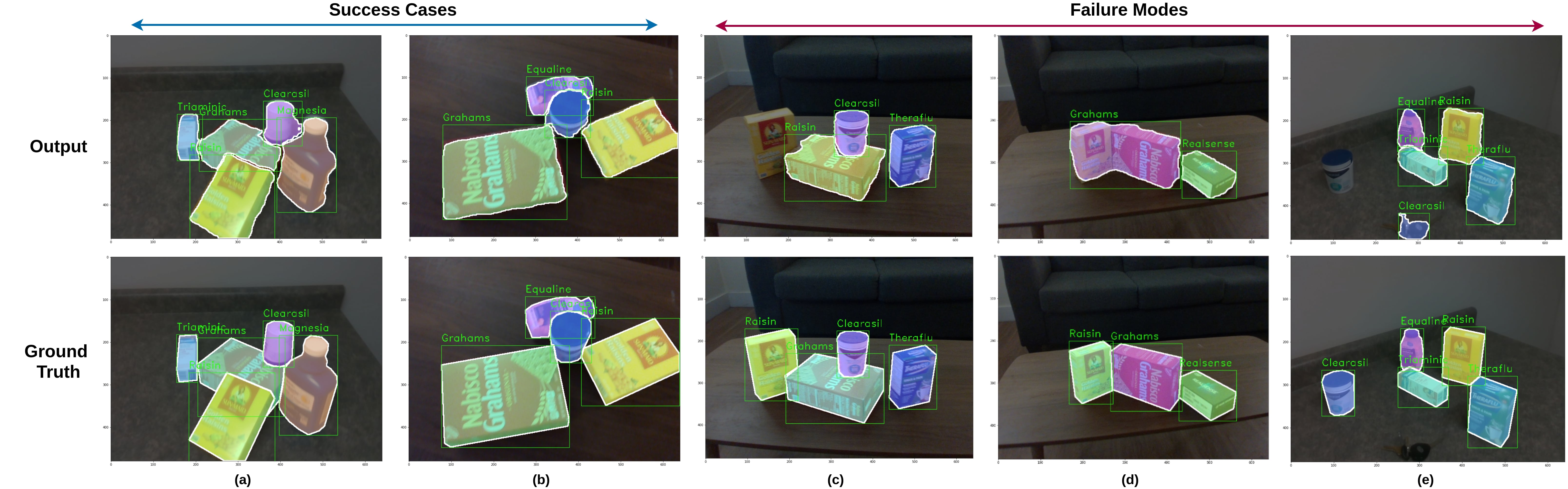}
\caption{\textbf{Qualitative Results.} (a)(b) Successful cases. Note the successful registration and labelling of highly occluded instances. (c) Instance registration fails due to segmentation network not recognizing the leftmost instance, culminating in a false negative. (d) Instance registration fails due to segmentation network considering two instances as a single instance, culminating in a false positive. (e) Similarity matching confuses the labels of two registered instances in the foreground.   }\vspace{-0.2in}
\label{fig:qualitative}
\end{figure*}
\para{Qualitative Results.} We describe some qualitative results in Fig \ref{fig:qualitative}. The two leftmost columns are the outputs of the pipeline at some instant along the image sequence for a successful stacked case in (a) and touching case in (b). Note that these cases are particularly challenging as both contain small objects that are significantly occluded. However, the pipeline is able to register and match instances in such occluded and cluttered scenes by combining information successively from multiple views, which allows the pipeline to be resilient to occlusion.

However, there are instances where the pipeline can fail. The causes of such failures can broadly be classified into two categories depending on which part of the pipeline caused the failure. Firstly, we have errors originating from the instance registration stage due to excessively noisy segmentation network outputs which cannot be rectified by our methods. We describe some of these cases in parts (c)(d). Secondly, we have errors originating from the similarity matching step. We find that such errors are far more rare compared to the first kind. We show one such case in part (e). 

\para{Scene-Level Metrics.} We present the scene level metrics across the three clutter levels in Table \ref{tab:scene_results}. Clearly, the pipeline is most effective when objects are clearly separated. In fact, in such cases, we find near perfect instance recall and matching precision. Surprisingly however, we found that both instance recall and matching precision for the touching cases are less than that for the stacked cases. Upon closer inspection, we found that this is due to the instance segmentation network having difficulty telling whether or not two laterally adjacent cuboidal objects belong to separate instances (see Fig \ref{fig:qualitative}d). This under-segmentation is explained in \cite{xiang_learning_2021,xie_unseen_2020} as a consequence of the network relying primarily on depth information due to being trained on non-photorealistic data, meaning that it cannot use RGB features to distinguish two adjacent cuboidal instances apart. By contrast, in the less structured stacked cases, stacked cuboidal objects are rarely perfectly aligned, allowing the network to distinguish them reliably.

\para{Frame-Level Metrics.} We observe a similar trend when projecting from the reconstructed scene into the output images (see Table~\ref{tab:frame_results}). In particular, we find that the free case offers the best results in terms of mask overlap, boundary overlap, and recall with respect to the ground truth. We also find that the pipeline performs worse in terms of recall in the touching cases when compared to the stacked cases, but slightly better in terms of mask overlap with the ground truth. This is a consequence of the scene-level instance recall for the touching cases being worse than that of the stacked cases due to inherent limitations with the segmentation network described previously. However, there is better performance in terms of mask overlap for the touching cases as the stacked cases frequently feature difficult angles with multiple occlusions, which tends to produce artifacts in the output of the segmentation network. Finally, we conduct an ablation study by removing the influence of the weighting function in Eq. \ref{eq:reward}. We can see that the overall performance is markedly degraded, especially for the more challenging touching and stacked cases.

\begin{table}
\centering
\begin{tabular}{|l|c|c|c|}
    \hline
    Metric & Free & Touching & Stacked\\
    \hline \hline 
    Instance Recall          & 0.96 & 0.86 & 0.91  \\
    Matching Precision     & 1.00 & 0.80 & 0.88 \\
    \hline
\end{tabular}
\caption{\textbf{Scene-Level Metrics.} Note near-perfect recall and precision in the free case, and performance when stacked is slightly better than when touching.}
\label{tab:scene_results}
\end{table}

\begin{table}
\centering
\begin{tabular}{|l|c|c|c|c|c|c|c|}
    \hline
    \multirow{2}{*}{Condition} & \multicolumn{3}{|c|}{Mask Overlap} &  \multicolumn{3}{|c|}{Boundary Overlap} & \multirow{2}{*}{Recall@.75}\\
    \cline{2-7}
    & F & P & R & F & P & R & \\
    \hline \hline 
    Free     & 0.88 & 0.94 & 0.86 & 0.67 & 0.71 & 0.66 & 0.86\\
    Touching & 0.78 & 0.84 & 0.75 & 0.49 & 0.53 & 0.48 & 0.62\\
    Stacked  & 0.75 & 0.83 & 0.71 & 0.51 & 0.55 & 0.49 & 0.77\\
    \hline
\end{tabular}\vspace{0.5mm}
\begin{tabular}{|l|c|c|c|c|c|c|c|}
    \hline
    \multirow{2}{*}{Condition} & \multicolumn{3}{|c|}{Mask Overlap} &  \multicolumn{3}{|c|}{Boundary Overlap} & \multirow{2}{*}{Recall@.75}\\
    \cline{2-7}
    & F & P & R & F & P & R & \\
    \hline \hline 
    Free     & 0.84 & 0.96 & 0.79 & 0.65 & 0.76 & 0.62 & 0.80\\
    Touching & 0.68 & 0.88 & 0.63 & 0.42 & 0.58 & 0.39 & 0.49\\
    Stacked  & 0.50 & 0.63 & 0.43 & 0.36 & 0.45 & 0.32 & 0.50\\
    \hline
\end{tabular}
\caption{\textbf{Frame-Level Metrics.} Top: Using weighting function $w(t)$. Bottom: Without using weighting function. }
\label{tab:frame_results}
\end{table}

\section{Related Work}

\para{Image Annotation in Robotic Environments.} Real world image annotation is a highly human labor intensive process. As a result, data annotation becomes the primary limiting factor in the generation of large scale datasets required for data-driven approaches to robotic perception, culminating in the rapid growth of the data annotation industry \cite{data_annotation}. Several tools and pipelines have been introduced to streamline the image annotation process as much as possible. LabelFusion \cite{marion_labelfusion_2017} is a pipeline which allows users to easily align known CAD models into cluttered scenes to generate labelled views of the scene from multiple viewing angles. EasyLabel \cite{suchi_easylabel_2019} follows an incremental approach, generating pixelwise annotations of cluttered scenes progressively as more items are added to the scene. However, to our knowledge no existing tool or pipeline allows for fully autonomous image annotation without continuous human supervision.

\para{RFID Tracking and Localization.} RFIDs are inexpensive battery-free tags commonly used in logistical applications such as the tracking of goods in warehouses. Their basic principle of operation is to use backscatter modulation to echo a 96-bit electronic product code (EPC) in response to a query signal sent by an RFID reader. As multiple RFID tags can respond to each query at the same time, it is useful to be able to locate the source of each reply to localize the corresponding RFID. There is a considerable body of work in accomplishing this task at varying levels of accuracy using nothing more than commercially available commodity RFID readers. STPP \cite{shangguan_relative_2015} uses spatio-temporal phase profiles of the reflected signal from each RFID in order to ascertain their relative ordering in a highly restrictive conveyor belt setting. Tagoram \cite{yang_tagoram_2014} follows a similar setting, but uses an ensemble of antennas to track the movement of RFIDs over time in a fine-grained manner. MobiTagBot \cite{shangguan_design_2016} leverages these ideas to implement a mobile robot capable of telling if a linear sequence of RFID-tagged objects, such as on a bookshelf or rack, is mis-sorted. However, these approaches are unsuitable for our problem as we depart from such highly structured planar settings and instead explore unstructured 3D environments. Furthermore, the approach in Tagoram is impractical for our setting as it is difficult to fit multiple large RFID antennas on a mobile robot.

\section{Discussion \& Conclusion}

We present RF-Annotate, a first step toward automating dense pixelwise image annotation for object recognition and segmentation tasks in robotic environments by taking advantage of recent advances in wireless tracking technology. RF-Annotate leverages two complementary sensing modalities, geometric depth and UHF radio waves, in order to obtain supervision for a third (RGB). To accomplish this task, it combines learning-based approaches on the depth channel with classical signal processing approaches on radio waves.  Our experiments verify that RF-Annotate can robustly enable automated annotations in common tabletop environments. As for future work, we believe there are two fruitful directions. First, investigating how to generalize the scheme to leverage other technologies like BLE or UWB tags. Secondly, leveraging more sophisticated sensor setups like multiple antennas, or finding better signal-processing algorithms for the task.

\bibliographystyle{IEEEtran}
\bibliography{IEEEabrv,egbib}

\end{document}